\documentclass{article}

\usepackage{microtype}
\usepackage{graphicx}
\usepackage{subfigure}
\usepackage{booktabs} 

\usepackage{amsmath}
\usepackage{amssymb}
\usepackage{amsthm}
\newcommand{\E}{\mathbb{E}}
\newcommand{\R}{\mathbb{R}}

\DeclareMathOperator*{\argmin}{arg\,min}

\newtheorem{claim}{Claim}

\usepackage{hyperref}

\usepackage[accepted]{icml2019}

\icmltitlerunning{The equivalence between SVGD and BBVI}

\begin{document}

\twocolumn[
\icmltitle{The equivalence between Stein variational gradient descent \\and black-box variational inference}




\icmlsetsymbol{equal}{*}

\begin{icmlauthorlist}
\icmlauthor{Casey Chu}{stanford}
\icmlauthor{Kentaro Minami}{pfn}
\icmlauthor{Kenji Fukumizu}{pfn,ism}
\end{icmlauthorlist}

\icmlaffiliation{stanford}{Stanford University}
\icmlaffiliation{pfn}{Preferred Networks}
\icmlaffiliation{ism}{The Institute of Statistical Mathematics}

\icmlcorrespondingauthor{Casey Chu}{caseychu@stanford.edu}
\hyphenation{caseychu}

\icmlkeywords{Machine Learning, ICML}

\vskip 0.3in
]



\printAffiliationsAndNotice{\icmlEqualContribution} 



\begin{abstract}
We formalize an equivalence between two popular methods for Bayesian inference: Stein variational gradient descent (SVGD) and black-box variational inference (BBVI). In particular, we show that BBVI corresponds precisely to SVGD when the kernel is the neural tangent kernel. Furthermore, we interpret SVGD and BBVI as \emph{kernel gradient flows}; we do this by leveraging the recent perspective that views SVGD as a gradient flow in the space of probability distributions and showing that BBVI naturally motivates a Riemannian structure on that space. We observe that kernel gradient flow also describes dynamics found in the training of generative adversarial networks (GANs). This work thereby unifies several existing techniques in variational inference and generative modeling and identifies the kernel as a fundamental object governing the behavior of these algorithms, motivating deeper analysis of its properties.
\end{abstract}

\section{Introduction}
The goal of Bayesian inference is to compute the \emph{posterior} $P(x|z)$ over a variable of interest $x$. In principle, this posterior may be computed from the \emph{prior} $P(x)$ and the \emph{likelihood} $P(z|x)$ of observing data $z$, using the equation \begin{equation}
	p(x) := P(x|z) = \frac{P(z|x)P(x)}{\int P(z|x)P(x)\,dx}.
\end{equation} We denote the posterior as $p(x)$ for convenience of notation. Unfortunately, the integral in the denominator is usually intractable, which motivates \emph{variational inference} techniques, which approximate the true posterior $p(x)$ with an approximate posterior $q(x)$, often by minimizing the KL divergence $\mathrm{KL}(q(x) \,\|\, p(x))$. In this paper, we consider two popular variational inference techniques, black-box variational inference \cite{ranganath2014black} and Stein variational gradient descent \cite{liu2016stein}, and show that they are equivalent when viewed as instances of \emph{kernel gradient flow}. 

\section{Stein variational gradient descent}
Stein variational gradient descent \cite{liu2016stein}, or SVGD, is a technique for Bayesian inference that approximates the true posterior $p(x)$ with a set of particles $x_1, \ldots, x_n$.

In the continuous-time limit of small step size, each particle undergoes the update rule \begin{equation}
	\frac{dx_i}{dt} = \E_{y \sim q_t}[k(x_i,y)\nabla_y \log p(y) + \nabla_y k(x_i,y)], \label{eq:svgd}
\end{equation} where $q_t$ denotes the empirical distribution of particles at time $t$: \begin{equation}
	q_t = \frac{1}{n}\sum_{i=1}^n  \delta_{x_i(t)},
\end{equation} and  $k(x,y)$ is a user-specified kernel function, such as the RBF kernel $k(x,y) = e^{-||x-y||^2}$.

In the mean-field limit as $n \to \infty$ \cite{lu2019scaling}, an equivalent form of the dynamics \eqref{eq:svgd} is obtained by an application of Stein's identity (integration by parts on the second term): \begin{equation}
	\frac{dx}{dt} = \E_{y \sim q_t}[k(x,y)\nabla_y (\log p(y) - \log q_t(y))].  \label{eq:svgd2}
\end{equation} 


\section{Black-box variational inference}

Black-box variational inference \cite{ranganath2014black}, or BBVI, is another technique for Bayesian inference that approximates the true posterior $p(x)$ with an approximate posterior $q_\phi(x)$, where $q_\phi$ is a family of distributions parameterized by $\phi$. In BBVI, we maximize the evidence lower bound, or ELBO, objective
\begin{equation}
	L(\phi) := \E_{x \sim q_\phi} \Big[ \log \frac{P(z|x)P(x)}{q_\phi(x)} \Big]
\end{equation}
by gradient ascent on $\phi$. This procedure effectively minimizes the KL divergence between $q_\phi(x)$ and the true posterior $p(x) = P(x|z)$, since the KL divergence and the ELBO objective differ by only the \emph{evidence} $P(z)$, which is constant w.r.t.~$\phi$: \begin{equation}
	\mathrm{KL}(q_\phi(x) \,\|\, p(x)) = P(z) - L(\phi). \label{eq:elbodiffersbyconstant}
\end{equation}

Our claim is:
\begin{claim}
The sequence of approximate posteriors generated by BBVI, when the reparameterization trick of \citet{kingma2013auto} is used, is governed by the SVGD dynamics \eqref{eq:svgd2}, where the kernel $k$ is the neural tangent kernel of \citet{jacot2018neural}.
\end{claim}

To see this, we observe that the evolution of the parameters $\phi$ under gradient ascent is governed by 
\begin{equation}
	\frac{d\phi}{dt} = \nabla_\phi L(\phi). \label{eq:bbvi}
\end{equation}

Next, we specialize to the case where the family of approximate posteriors is parameterized via the reparameterization trick \cite{kingma2013auto}. That is, suppose that there exists a fixed distribution $\omega$ and a parameterized function $f_\phi$ such that the following two sampling methods result in the same distribution over $x$: \begin{equation}
	x \sim q_\phi \iff \varepsilon \sim \omega \text{ and } x = f_\phi(\varepsilon).
\end{equation} As an example, the family of normal distributions $\mathcal{N}(\mu, \sigma)$ may be reparameterized as \begin{equation}
	x \sim \mathcal{N}(\mu, \sigma) \iff \varepsilon \sim \mathcal{N}(0,1) \text{ and } x = \mu + \sigma\varepsilon.
\end{equation}
In this setting, \citet{roeder2017sticking} and \citet{feng2017learning} noted that
\begin{equation}
\begin{aligned}
	&\nabla_\phi L(\phi) \\
	&\qquad= \E_{w \sim \omega} [ \nabla_\phi f_\phi(w) \cdot \nabla_y (\log p(y) - \log q_\phi(y) )\big|_{y = f_\phi(w)}  ].
\end{aligned} \label{eq:stickingthelanding}
\end{equation}

Now, we consider the dynamics of a sample $x = f_\phi(\varepsilon)$ under the parameter dynamics \eqref{eq:bbvi}. By the chain rule, we have that \begin{equation}
	\frac{dx}{dt} = (\nabla_\phi f_\phi(\varepsilon))^T \frac{d\phi}{dt}. \label{eq:chainrule}
\end{equation} 
Let us introduce the \emph{neural tangent kernel} of \citet{jacot2018neural} \begin{equation}
	\Theta_\phi(\varepsilon, w) := (\nabla_\phi f_\phi(\varepsilon))^T \nabla_\phi f_\phi(w),
\end{equation} and define 
\begin{equation}
	k_\phi(x, y) := \Theta_\phi(f^{-1}_\phi(x), f^{-1}_\phi(y)), \label{eq:ntkkernel}
\end{equation} making the additional assumption that $\varepsilon \mapsto f_\phi(\varepsilon)$ is injective. Note that if $x \in \R^n$, then $\Theta_\phi(\varepsilon, w)$ and $k_\phi(x, y)$ are both $n$-by-$n$ matrices that depend on $\phi$. Then, substituting \eqref{eq:bbvi} and \eqref{eq:stickingthelanding} into \eqref{eq:chainrule}, we find that the samples satisfy \begin{align}
	\frac{dx}{dt} &= (\nabla_\phi f_\phi(\varepsilon))^T \frac{d\phi}{dt} \\
	&= \E_{w \sim \omega} [ \Theta_\phi(\varepsilon, w)\, \nabla_y (\log p(y) - \log q_\phi(y) )\big|_{y = f_\phi(w)}  ] \\
	&= \E_{y \sim q_\phi} [ k_{\phi}(x, y)\, \nabla_y (\log p(y) - \log q_\phi(y) )  ]. \label{eq:bbvikgf}
\end{align} 

Comparing \eqref{eq:bbvikgf} with the SVGD dynamics \eqref{eq:svgd2}, we find an exact correspondence between SVGD and BBVI, where in BBVI, the kernel is given by \eqref{eq:ntkkernel} and defined by the neural tangent kernel.

\subsection{Example: a Gaussian variational family}
\begin{figure*}
	\centering
	\includegraphics[height=2in]{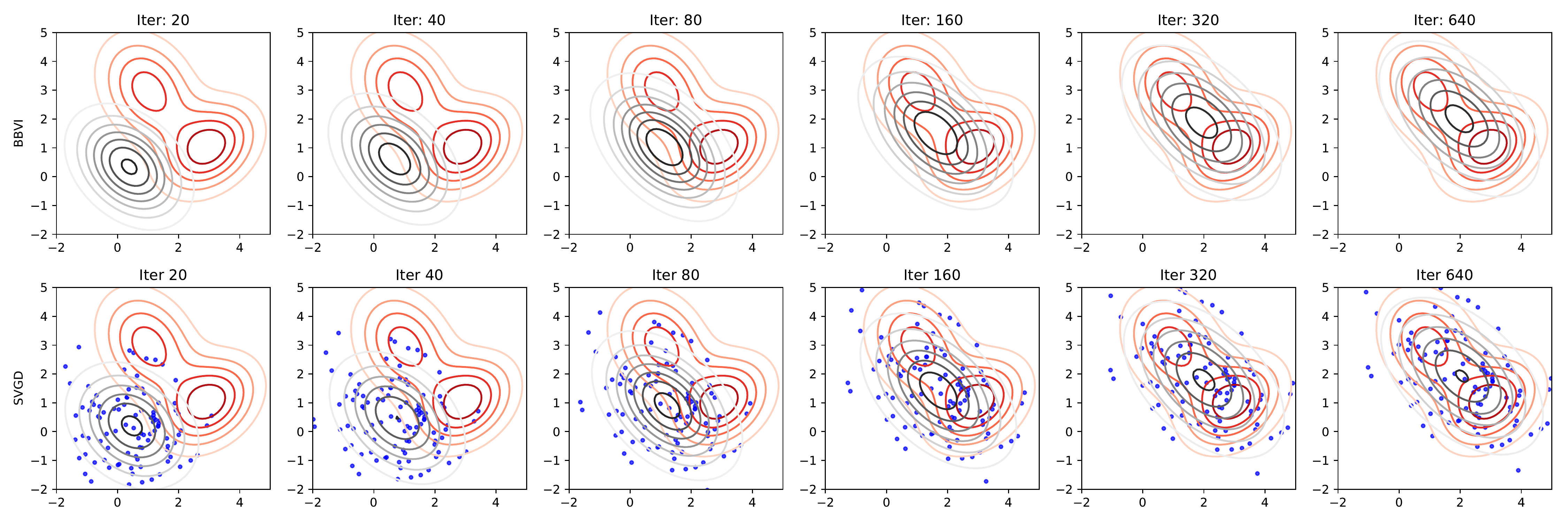}
	\caption{The sequence of approximate posteriors obtained by BBVI and SVGD with the theoretically equivalent kernel.} \label{fig:bbvisvgd}
\end{figure*}

As an example, consider the family of multivariate normal distributions $\mathcal{N}(\mu, \Sigma)$, parameterized by an invertible matrix $A$ and a vector $\mu$, with the relation $\Sigma = AA^T$. This variational family is reparameterizable with \begin{equation}
	x \sim \mathcal{N}(\mu, \Sigma) \iff \varepsilon \sim \mathcal{N}(0, I) \text{ and } x = \mu + A\varepsilon. \label{eq:gaussianreparam}
\end{equation}
In this setting, the kernel \eqref{eq:ntkkernel} becomes \begin{equation}
	k(x, y) =  (1 + (x - \mu)^T \Sigma^{-1} (y - \mu)) I, \label{eq:gaussianbbvikernel}
\end{equation} where $I$ is the identity matrix. In the continuous-time and many-particle limit, BBVI with the parameterization \eqref{eq:gaussianreparam} produces the same sequence of approximate posteriors as SVGD with the kernel \eqref{eq:gaussianbbvikernel}. \autoref{fig:bbvisvgd} compares the sequence of approximate posteriors generated by BBVI and SVGD with the theoretically equivalent kernel \eqref{eq:gaussianbbvikernel} in fitting a bimodal 2D distribution; we see that the agreement is quite close.

It is instructive to perform the computation of \eqref{eq:gaussianbbvikernel} explicitly. We use index notation with Einstein summation notation, where indices that appear twice are implicitly summed over. We have that $f_i(\varepsilon) = \mu_i + A_{ik} \varepsilon_k$ and \begin{equation}
	\frac{\partial f_i(\varepsilon)}{\partial \mu_\ell} =\delta_{i\ell}, \quad 
	\frac{\partial f_i(\varepsilon)}{\partial A_{\ell m}} = \delta_{i\ell}\delta_{km} \varepsilon_k,
\end{equation}
so that the neural tangent kernel is
\begin{align}
	\Theta_{ij}(\varepsilon, w) 
		&= \frac{\partial f_i(\varepsilon)}{\partial \mu_\ell} \frac{\partial f_j(w)}{\partial \mu_\ell} + \frac{\partial f_i(\varepsilon)}{\partial A_{\ell m}} \frac{\partial f_j(w)}{\partial A_{\ell m}} \\
		&= \delta_{i\ell}\delta_{j\ell} + \delta_{i\ell}\delta_{km} \varepsilon_k \delta_{j\ell}\delta_{om} w_o \\
		&= \delta_{ij} +  \delta_{ij} \varepsilon_m w_m,
\end{align}
or $\Theta(\varepsilon, w) = (1 + \varepsilon \cdot w)I$ in vector notation. Then, using the definition \eqref{eq:ntkkernel} and substituting $f^{-1}(x) = A^{-1}(x - \mu)$ and $\Sigma = AA^T$, we arrive at \eqref{eq:gaussianbbvikernel}.

\section{Motivating a Riemannian structure}
In the previous section, we found that SVGD and BBVI both correspond to particle dynamics of the form
\begin{equation}
	\frac{dx}{dt} = \E_{y \sim q_\phi} [ k_{\phi}(x, y)\, \nabla_y (\log p(y) - \log q_\phi(y) )  ].
\end{equation}
One peculiar feature of the BBVI dynamics is that the kernel $k_\phi$ depends on the current parameter $\phi$, rather than being constant as the approximate posterior $q_\phi$ changes, as in the SVGD case. 

In fact, we argue that this feature of BBVI is quite natural:

\begin{claim}
	The requirement of BBVI that the kernel depends on the current distribution naturally motivates a Riemannian structure on the space of probability distributions.
\end{claim}

To make this claim, let us first review Euclidean and Riemannian gradient flows. In Euclidean space, following the negative gradient of a function $J : \R^n \to \R$ according to \begin{equation}
	\frac{dx}{dt} = -\nabla J(x)  \label{eq:euclideangradientflow}
\end{equation} can lead to a minimizer of $J$. Analogously, on a \emph{Riemannian manifold} $M$, following the negative \emph{Riemannian gradient} of a function $J : M \to \R$ according to \begin{equation}
	\frac{dx}{dt} = -G(x)^{-1} \nabla J(x), \label{eq:riemanniangradientflow}
\end{equation} can lead to a minimizer of $J$. Here, $G$ is a positive-definite matrix-valued function called the \emph{Riemannian metric}, which defines the local geometry at $x$ and perturbs the Euclidean gradient $\nabla J$ pointwise. Note that in the case that $G(x)$ is the identity matrix for all $x$, Riemannian gradient flow reduces to the Euclidean gradient flow.

Next, we review Wasserstein gradient flows, which generalize gradient flows to the space of probability distributions \cite{ambrosio2008gradient}. Here, we consider the set of all probability distributions over a particular space formally as an ``infinite-dimensional'' manifold $\mathcal{P}$, and we consider a function $J : \mathcal{P} \to \R$. In variational inference, the most relevant such function is the KL divergence $J(q) := \mathrm{KL}(q \,\|\, p)$, where we are interested in finding an approximate posterior that minimizes $J$. Analogous to before, a minimizer of $J$ may be obtained by following the analogue of a gradient; the trajectory of the distribution $q$ turns out to take the form of the PDE \begin{equation}
	\frac{\partial q}{\partial t} = \nabla \cdot (q \nabla \Psi_q). \label{eq:wassersteingradientflow}
\end{equation} Here, $\nabla \Psi_q$ serves as the correct analogue of the gradient of $J$ evaluated at $q$, and it turns out that $\Psi_q(x) = \log q(x) - \log p(x)$ for the variational inference case $J(q) := \mathrm{KL}(q \,\|\, p)$. This function $\Psi_q$ is known variously as the functional derivative, first variation, or von Mises influence function.

Now, we review the recent perspective that SVGD can be interpreted as a generalized Wasserstein gradient flow under the \emph{Stein geometry} \cite{liu2017stein,pmlr-v97-liu19i,duncan2019geometry}. We follow the presentation of \citet{duncan2019geometry} and refer to it for a rigorous treatment. To set the stage, we take a non-parametric view of the SVGD update \eqref{eq:svgd}, in which the dependence on $\phi$ is interpreted as dependence on the distribution $q$ itself: \begin{equation}
	\frac{dx}{dt} = \E_{y \sim q} [ k(x, y)\, \nabla_y (\log p(y) - \log q(y) )  ].
\end{equation} Substituting $\Psi_q(y) = \log q(y) - \log p(y)$ and the linear operator $T_{q}$ defined by \begin{equation}
	(T_{q} \varphi)(x) := \E_{y \sim q}[k(x,y) \varphi(y)], \label{eq:kerneloperator}
\end{equation} we have \begin{equation}
	\frac{dx}{dt} = -(T_q \nabla \Psi_q)(x).
\end{equation}
Under these dynamics, the probability distribution $q$ evolves according to the PDE \begin{equation}
	\frac{\partial q}{\partial t} = \nabla \cdot (q T_{q} \nabla \Psi_q). \label{eq:steingradientflow}
\end{equation} 
Comparing \eqref{eq:steingradientflow} with \eqref{eq:wassersteingradientflow}, we see that \eqref{eq:steingradientflow} defines a modified gradient flow in which the gradient $\nabla \Psi_q$ is perturbed by the operator $T_q$. 

We now advocate for generalizing Wasserstein gradient flow in the same way that Riemannian gradient flow generalizes Euclidean gradient flow. The operator $T_q$ perturbs the gradient in a way analogous to how the Riemannian metric perturbs the Euclidean gradient in \eqref{eq:riemanniangradientflow}, so the operator $T_q$ thereby defines an analogue of a Riemannian metric on $\mathcal{P}$. However, there is no fundamental reason that $T_q$ must have the restrictive form prescribed by \eqref{eq:kerneloperator}. Indeed, because $T_q$ is analogous to the Riemannian metric $G(x)$, it is natural to let the kernel, whose action defines the operator $T_q$, depend on the current value of $q$. It is also natural to allow the kernel to output a matrix rather than a scalar so that $T_q$ may mix all components of $\varphi$. \citet{duncan2019geometry} in fact speculate on these possibilities (Remarks 17 and 1).

With these considerations in mind, we propose replacing \eqref{eq:kerneloperator} with \begin{equation}
	(T_q \varphi)(x) := \E_{y \sim q}[k_q(x,y) \varphi(y)], \label{eq:kerneloperator2}
\end{equation} where the kernel $k_q$ now depends on $q$ and outputs a matrix. This defines a gradient flow by \eqref{eq:steingradientflow} that we will refer to as \emph{kernel gradient flow}.\footnote{To further the analogy between Euclidean and Riemannian gradient flow and Wasserstein and kernel gradient flow, note that just as setting the Riemannian metric to identity matrix for all $x$ reduces Riemannian to Euclidean gradient flow, setting $T_q$ to the identity operator for all $q$ reduces kernel to Wasserstein gradient flow. The special ``Euclidean'' Riemannian metric obtained this way is the central object of the \emph{Otto calculus} \cite{otto2001geometry,ambrosio2008gradient}.}

Once $T_q$ has the form \eqref{eq:kerneloperator2}, BBVI may naturally be regarded as an instance of kernel gradient flow, in which the kernel $k_q$ is the neural tangent kernel which depends on the current distribution $q$. More abstractly, we see that the neural tangent kernel defines a Riemannian metric on the space of probability distributions. We summarize the perspective that this framework gives on variational inference:
\begin{claim}
	SVGD updates generate a kernel gradient flow of the loss function $J(q) := \mathrm{KL}(q\,\|\,p)$, with a Riemannian metric determined by the user-specified kernel.
\end{claim}
\begin{claim}
	BBVI updates generate a kernel gradient flow of the loss function $J(q) := \mathrm{KL}(q\,\|\,p)$, with a Riemannian metric determined by the neural tangent kernel of $f_\phi$.
\end{claim}

\section{Beyond variational inference: GANs as kernel gradient flow}

We now argue that the kernel gradient flow perspective we have developed describes not only SVGD and BBVI, but also describes the training dynamics of \emph{generative adversarial networks} \cite{goodfellow2014generative}.

Generative adversarial networks, or GANs, are a technique for learning a generator distribution $q_\phi$ that mimics an empirical data distribution $p_{\mathrm{data}}$. The generator distribution $q_\phi$ is defined implicitly as the distribution obtained by sampling from a fixed distribution $\omega$, often a standard normal, and running the sample through a neural network $f_\phi$ called the \emph{generator}. The learning process is facilitated by another neural network $D_\theta$ called the \emph{discriminator} that takes a sample and outputs a real number, and is trained to distinguish between a real sample from $p_{\mathrm{data}}$ and a fake sample from $q_\phi$. The generator and discriminator are trained simultaneously until the discriminator is unable to distinguish between real and fake samples, at which point the generator distribution $q_\phi$ hopefully mimics the data distribution $p_{\mathrm{data}}$.

For many GAN variants, the rule to update the generator parameters $\phi$ can be expressed in the continuous-time limit as \begin{equation}
    \frac{d\phi}{dt} = \nabla_\phi \E_{w \sim \omega} [ D_\theta(f_\phi(w))], \label{eq:gan}
\end{equation} or by the chain rule, 
\begin{equation}
    \frac{d\phi}{dt} = \E_{w \sim \omega} [ \nabla_\phi f_\phi(w) \cdot \nabla_y D_\theta(y)\big|_{y = f_\phi(w)}  ].
\end{equation}
The discriminator parameters $\theta$ are updated simultaneously to minimize a separate discriminator loss $L_\text{disc}(\theta, \phi)$, but it is common for theoretical purposes to assume that the discriminator achieves optimality at every training step. Denoting this optimal discriminator as $-\Psi_\phi$ (i.e.~setting $\Psi_\phi := -D_{\theta^*}$ for $\theta^* := \argmin_\theta L_\text{disc}(\theta, \phi)$), we have 
\begin{equation}
    \frac{d\phi}{dt} = -\E_{w \sim \omega} [ \nabla_\phi f_\phi(w) \cdot \nabla_y \Psi_\phi(y)\big|_{y = f_\phi(w)}  ].
\end{equation} This matches the BBVI update rule \eqref{eq:stickingthelanding} with $\log q_\phi(y) - \log p(y)$ replaced by the discriminator $\Psi_\phi(y)$. Hence, analogous to \eqref{eq:bbvikgf}, the generated points $x$ satisfy
\begin{equation}
	\frac{dx}{dt} = -\E_{y \sim q_\phi} [ k_{\phi}(x, y)\, \nabla_y \Psi_\phi(y)  ], \label{eq:gankgf}
\end{equation} where here $k_\phi$ is defined as in \eqref{eq:ntkkernel} by the neural tangent kernel of the generator $f_\phi$. Finally, it was observed that the optimal discriminator $\Psi_q$ of the minimax GAN equals the functional derivative of the Jensen--Shannon divergence \cite{chu2019probability,chu2020smoothness}; hence we conclude:

\begin{claim}
	Minimax GAN updates generate a kernel gradient flow of the Jensen--Shannon divergence  $J(q) := \mathrm{D}_{\mathrm{JS}}(p_{\text{data}}, q)$, with a Riemannian metric determined by the neural tangent kernel of the generator $f_\phi$.
\end{claim}
 Similarly, non-saturating and Wasserstein GAN updates generate kernel gradient flows on the directed divergence $J(q) := \mathrm{KL}(\frac{1}{2}p_{\text{data}} + \frac{1}{2}q \,\|\, p_{\text{data}})$ and Wasserstein-1 distance $J(q) := W_1(p_{\text{data}}, q)$ respectively.


\section{Conclusion}
We have cast SVGD and BBVI, as well as the dynamics of GANs, into the same theoretical framework of kernel gradient flow, thus identifying an area ripe for further study.

\bibliography{main}
\bibliographystyle{icml2019}

\end{document}